\documentclass{article}
\usepackage{spconf,amsmath,graphicx}

\usepackage[dvipsnames]{xcolor}

\usepackage{booktabs}

\usepackage{amssymb}
\usepackage{algorithm}
\usepackage[noend]{algpseudocode}
\usepackage{commath,url}
\usepackage{amsmath}

\usepackage{amsmath}
\DeclareMathOperator*{\argmax}{argmax} 
\usepackage{xparse}

\NewDocumentCommand{\tens}{e{_^}}{%
  \mathbin{\mathop{\otimes}\displaylimits
    \IfValueT{#1}{_{#1}}
    \IfValueT{#2}{^{#2}}
  }%
}
\newcommand{\eat}[1]{}

\title{ Recovery of Joint Probability Distribution from One-Way Marginals: Low Rank Tensors and Random Projections}

\name{Jian Vora\textsuperscript{1}, Karthik S. Gurumoorthy\textsuperscript{2}, Ajit Rajwade\textsuperscript{3}\thanks{Email addresses of the authors are 
{\texttt{jianvora@iitb.ac.in}}, {\texttt{gurumoor@amazon.com}}, and  
{\texttt{ajitvr@cse.iitb.ac.in}}. AR thanks SERB Matrics Grant \#10013890.}}
\address{Dept. of EE., IIT Bombay, India\textsuperscript{1};
India Machine Learning, Amazon, India\textsuperscript{2};\\
Dept. of CSE., IIT Bombay, India\textsuperscript{3}}
\def\BibTeX{{\rm B\kern-.05em{\sc i\kern-.025em b}\kern-.08em T\kern-.1667em\lower.7ex\hbox{E}\kern-.125emX}}
\begin{document}
%
\maketitle
\begin{abstract}
Joint probability mass function (PMF) estimation is a fundamental machine learning problem. The number of free parameters scales exponentially with respect to the number of random variables. Hence, most work on nonparametric PMF estimation is based on some structural assumptions such as clique factorization adopted by probabilistic graphical models, imposition of low rank on the joint probability tensor and reconstruction from 3-way or 2-way marginals, etc. In the present work, we link random projections of data to the problem of PMF estimation using ideas from tomography. We integrate this idea with the idea of low-rank tensor decomposition to show that we can estimate the joint density from just one-way marginals in a transformed space. We provide a novel algorithm for recovering factors of the tensor from one-way marginals, test it across a variety of synthetic and real-world datasets, and also perform MAP inference on the estimated model for classification.
\end{abstract}
\vspace{-0.2cm}
\section{Introduction}
\label{sec:intro}
Learning the joint distribution of $N$ random variables (RVs) is a fundamental problem in machine learning, signal processing and statistics \cite{murphy2012machine}, \cite{wainwright_2019}. More specifically, given $N$ RVs $\{X_1, X_2, .., X_N\}$, the goal is to estimate their joint probability density function (PDF) or their joint probability mass function (PMF) from their samples with potentially missing entries.\eat{ In the case of discrete RVs, we have
$p(i_1, i_2,..., i_N) = P(X_1=i_1, X_2 = i_2,..., X_N = i_N)$.} The joint PDF/PMF is used in many applications including, (1) as a generative model to produce new data-points by sampling,\eat{ i.e., $\boldsymbol{\hat{x}} \sim p(X_1, X_2,...,X_N)$,} (2) for conditional inferences as having the joint implies we can find any conditional PDFs/PMFs.\eat{ of the form $p(X_i,X_j,..X_k|X_p, X_q,...X_r)$\footnote{These however may not be possible to be computed exactly and tractably for a general density $p(X_1, X_2,...,X_N)$. There exist a class of models which allow for tractable inference including this work for discrete variables.}. As an example, consider we have labeled data in the form of tuples $(\boldsymbol{x_i},y_i)$ when $\boldsymbol{x_i}$ is a data-point and $y_i$ is the corresponding label. If we can estimate the joint PMF/PDF $p(\boldsymbol{x},y)$, then MAP inference used for classification is defined as $\hat{y} = \argmax_y p(y|\boldsymbol{x}) = \argmax_y p(\boldsymbol{x},y)/\sum_y p(\boldsymbol{x},y)$.} 
For discrete variables, the most common solution is to use empirical frequency of a tuple as a proxy for the joint distribution \cite{tsybakoc2008}. However, this simple method suffers from poor sample complexity $\gg \Omega(I^N)$  where $I$ is the number of distinct values that each variable $X_i$ can acquire. 
One popular route is to make use of problem-specific
structural assumptions such as in graphical models \cite{Jordan1999} to help reduce the model complexity. However, besides the graphical structure being unknown, inference in these models is difficult due to intractable posterior integrals, requiring randomization (e.g.\ MCMC \cite{mackay1998introduction}) or approximation (e.g.\ variational Bayes methods \cite{Jordan1999}). Recently, imposing a low rank assumption on the underlying probability tensor has gained popularity \cite{Kargas2018} where the $N$-way low-rank joint PMFs are estimated from only 3-way marginals. This is further extended to using only 2-way marginals in \cite{Ibrahim2020}.\\ \textbf{Our contributions:} In this work, we present a novel algorithm that combines the strengths of low-rank tensor approximation and techniques from computed tomography to obtain $N$-way joint PMFs from \emph{just one-way marginals}. 

\vspace{-0.2cm}
\section{Background}
\subsection{Canonical Polyadic Decomposition of PMF Tensors}
An $N-$way tensor $\mathcal{Z} \in \mathbb{R}^{I_1 \times I_2 \times I_3 \times ...\times I_N}$ representing the joint PMF of $N$ discrete RVs where $\mathcal{Z}(x_1,x_2,\ldots,x_N) = p(X_1=x_1,X_2=x_2,\ldots,X_N=x_N),$ admits a `Canonical Polyadic Decomposition' (CPD) if it can be decomposed as a sum of $F$ rank-$1$ tensors \cite{animacpd}. Denoting $a \tens b$ as the outer-product of two vectors, the CPD model is:
\vspace{-0.1in}
\begin{equation}
\mathcal{Z} = \sum_{f=1}^{f=F}\boldsymbol{\lambda}(f)\boldsymbol{A_1}(:,f) \tens \boldsymbol{A_2}(:,f) \tens ... \tens \boldsymbol{A_N}(:,f). 
\label{eq:tens_dec}
\end{equation}

Here $F$ is the smallest number for which such a decomposition exists and is called the rank of the tensor, $\boldsymbol{\lambda} \in \mathbb{R_+}^F$ and the mode latent factors $\boldsymbol{A_i} \in \mathbb{R_+}^{I_i \times F}$. The entries of $\boldsymbol{\lambda}$ and each column of the mode latent factor are non-negative and sum to one. Recovering the PMF tensor $\mathcal{Z}$ is equivalent of estimating the factors $\boldsymbol{\lambda},\boldsymbol{A_1},\boldsymbol{A_2},...,\boldsymbol{A_N}$ \cite{sidirotens}.

As explained in \cite{Kargas2018}, the CPD model also has a probabilistic interpretation of being a naive Bayes model by introducing an additional latent variable $H$ with $P(H=f) =  \boldsymbol{\lambda}(f)$ and the mode latent factors are conditional densities given $H$.

Assume a latent variable $H$ taking $F$ distinct states, then the same CPD model as described above can be formulated in the following manner:

\begin{multline}
     \mathcal{Z}  = \sum_{f=1}^{f=F} Pr(H=f) \prod_{n=1}^{n=N}Pr(X_n=i_n|H=f)
\end{multline}

Thus the mode latent factors have an interpretation of conditional densities. Thus, comparing with the original CPD model, we get $\boldsymbol{\lambda}$($f$) = $Pr(H=f)$ and $\boldsymbol{A_n}(i_n,f) = Pr(X_n=i_n|H=f)$. As it is clearly evident, the entries of $\boldsymbol{\lambda}$ and each column of the mode latent factor should sum to one to be valid densities along with the non-negativity constraints. $F$ is an important hyperparameter which explores how much dependency between the variables would we like to model. This is the decomposition which we shall use for this work which has been shown to be expressive enough for modelling a variety of densities in \cite{Kargas2018}.

\vspace{-0.2cm}
\subsection{Joint PMF Estimation from 3-way or 2-way Marginals}
\label{subsec:3way-2way}
The work in \cite{Kargas2018} showed the benefits of modelling joint PMFs as low-rank tensors via the CPD decomposition. The low tensor rank is motivated by the fact that in many real world datasets, the RVs are not fully dependent (nor fully independent). They proposed recovering $N$-way joint PMFs from 3-way marginals of the form  $\boldsymbol{Z}_{i,j,k} \triangleq p(X_i,X_j,X_k)$ which are obtained from the data using standard histogramming. From these marginals, the latent mode factors are recovered from the following coupled tensor factorization (CTF) estimator:
\begin{eqnarray}
\min_{\{\boldsymbol{A_n}\}_{n=1}^N,\boldsymbol{\lambda}} \sum_{j} \sum_{k>j} \sum_{l>k} \| \boldsymbol{Z}_{j,k,l} -  [\boldsymbol{\lambda},\boldsymbol{A_j},\boldsymbol{A_k},\boldsymbol{A_l}] \|_F^2,  \nonumber \\
\textrm{ s.t. } \forall n,f, \|\boldsymbol{A_n}(:,f)\|_1 =   \boldsymbol{\|\lambda}\|_1 = 1,
\boldsymbol{A_n} \succeq \boldsymbol{0};
\boldsymbol{\lambda} \succeq \boldsymbol{0}, \nonumber
\end{eqnarray}
where $\succeq$ represents the element-wise\eat{, $\geq$ inequality} and  $[\boldsymbol{\lambda},\boldsymbol{A_j},\boldsymbol{A_k},\boldsymbol{A_l}] \triangleq \sum_{f=1}^F \boldsymbol{\lambda}(f) \boldsymbol{A_j}(:,f) \tens \boldsymbol{A_k}(:,f) \tens \boldsymbol{A_l}(:,f)$. The problem with this approach is two fold: (a) for large $N$, the number of 3-way marginals increases rapidly as $O(N^3)$, and (b) accurately estimating the 3-way marginals is difficult \eat{in the common cases}when the number of samples is small or if the samples have missing entries.

The work in \cite{Ibrahim2020} recovers the joint PMF from only 2-way marginals $\boldsymbol{Z}_{j,k} = p(X_j,X_k)$ computed from the samples via histogramming. The mode latent factors are obtained using the relation $\boldsymbol{Z}_{j,k} = \boldsymbol{A_j} D(\boldsymbol{\lambda}) \boldsymbol{A_k}^T$ where $D(\boldsymbol{\lambda})$ is a diagonal matrix with $\boldsymbol{\lambda}$ on its diagonal. As the tensor rank $F \gg \text{min }(I_j, I_k)$ \cite{FuNMF}, methods like non-negative matrix factorization (NMF) \cite{Lee2000} cannot be directly applied to $\boldsymbol{Z}_{j,k}$ to determine $\boldsymbol{A_j}$ and $\boldsymbol{A_k}$. Instead, the work in \cite{Ibrahim2020} splits the indices of $N$ variables into sets $\mathcal{S}_1 = \{l_1, l_2, ...,l_M\}$ and $\mathcal{S}_2 = \{l_{M+1}, l_{M+2}, ...,l_N\}$ and constructs the matrix  $\boldsymbol{\widetilde{Z}}$ by row and column-concatenation of $\{\boldsymbol{Z}_{j,k}\}$ (see \cite[Eqn. 3]{Ibrahim2020}) from indices in $\mathcal{S}_1, \mathcal{S}_2$. Then, $\boldsymbol{\widetilde{Z}}$ is then decomposed as $\boldsymbol{\widetilde{Z}} = \boldsymbol{WH}^T$ using the successive projection algorithm (\textsc{Spa}) based on NMF literature \cite{Gillis2014}. The mode factors are extracted from $\boldsymbol{W}$ and $\boldsymbol{H}$ using the relation $\boldsymbol{W}= [\boldsymbol{A_{l_1}}, \boldsymbol{A_{l_2}},...,\boldsymbol{A_{l_M}}]^T$ and $\boldsymbol{H}^T= D(\boldsymbol{\lambda})[\boldsymbol{A_{l_{M+1}}}, \boldsymbol{A_{l_{M+2}}},..,\boldsymbol{A_{l_N}}]$. 
The formulation of the CPD model is similar to that of a Gaussian mixture model (GMM) where instead of the mixture weights, we have $\boldsymbol{\lambda}$ and instead of estimating the means and covariances of the Gaussians, we have to estimate the mode latent factors. Thus, the work in \cite{Yeredor2019MaximumLE} tries to find the parameters by maximising the likelihood of observing the data. It uses an expectation maximizaton (refered to as \textsc{Rand-em} henceforth) update like the one used for GMM training.
\[  \text{log }P(y[1],y[2],...,y[T]) = \sum_{t=1}^{t=T}\text{log }\sum_{f=1}^{f=F} \lambda_f \prod_{n=1}^{n=N}\boldsymbol{A_n}(y[t],f)    \]
Thus, they formulate the following optimization problem for minimizing the negative log-likelihood of having observed samples $y_t$:
\begin{equation}
\min_{\{\boldsymbol{A_n}\}_{n=1}^N,\boldsymbol{\lambda}} -\sum_{t=1}^{t=T}\text{log }\sum_{f=1}^{f=F} \lambda_f \prod_{n=1}^{n=N}\boldsymbol{A_n}(y_t,f) \end{equation}
A few expectation maximization (EM) iterations are subsequently executed using the output of the \textsc{Spa} as the initial condition \cite{Yeredor2019MaximumLE}, to further boost the performance.
\vspace{-0.3cm}
\subsection{Random Projections and PMF Estimation}
In this subsection (and further in Sec. \ref{sec:prob_stmt}) we show the link between random projections, tomography, and tensor recovery.
\textbf{Radon Transform:} The Radon Transform of a $N$-D function $p(\boldsymbol{x}):\mathbb{R}^N \to \mathbb{R}$, including the case where $p(.)$ is  the joint PDF of $\boldsymbol{x}$, in a direction $\boldsymbol{\phi}$ ($\|\boldsymbol{\phi}\|_2 = 1$) is defined as follows:
\begin{equation}
 \mathfrak{R}_{\boldsymbol{\phi}}(p)(t) = \int p(\boldsymbol{x})\delta(t - \boldsymbol{\phi}^t \boldsymbol{x})d\boldsymbol{x},
 \label{eq:radon}
 \end{equation}
where $\delta$ is the Dirac delta function and $t \in \mathbb{R}$ is an offset. The Radon Transform can be inverted to reconstruct $p(\boldsymbol{x})$ using methods like filtered back projection (FBP) \cite{Kak2001}, which are computationally efficient in 2D or 3D.\eat{ The discrete version of the Radon Transform is defined similarly by replacing the integral with a summation.}

We are interested in estimating $p(\boldsymbol{x})$ from random linear projections of the form $\boldsymbol{y} = \boldsymbol{\Phi x}$
where $\boldsymbol{\Phi} \in \mathbb{R}^{M \times N}$ with entries drawn i.i.d. from $\mathcal{N}(0,1)$ and then normalized so that each row has unit magnitude. We demonstrate that such data transformation aids in PDF/PMF estimation. For this, consider a vector $\boldsymbol{x} \in \mathbb{R}^N$ and a row $\boldsymbol{\phi}$ of $\boldsymbol{\Phi}$. The 1D PMF of the projection $\boldsymbol{\phi}^t \boldsymbol{x}$ is:
\eat{
\begin{multline}
p(\boldsymbol{\phi}^t \boldsymbol{x} = t) =p(\sum_{i=1}^N \phi_i x_i = t)=\sum_{a_1,a_2,...,a_{N-1}}p(\phi_1 x_1 = a_1,\\\phi_2 x_2 = a_2,..., \phi_{N-1}x_{N-1} = a_{N-1},\phi_N x_N = t - \sum_{i=1}^{i=N-1}a_i)
\label{eq:radon2}
\end{multline}
}
\vspace{-0.3cm}
\begin{equation}
p(\boldsymbol{\phi}^t \boldsymbol{x} = t) =  \sum_{\boldsymbol{x}}p(\boldsymbol{x})\delta_{kron}\left(\boldsymbol{\phi}^t \boldsymbol{x}-t\right).
\label{eq:radon2}
\end{equation}
Comparing Eqns. \ref{eq:radon} and \ref{eq:radon2}, we note that the PMF of $\boldsymbol{\phi}^t \boldsymbol{x}$ is in fact the (discretized) Radon Transform of the joint PMF tensor taken in the direction $\boldsymbol{\phi}$. Thus, if we collect 1D PMFs of the form $p(\boldsymbol{\phi_m}^t \boldsymbol{x})$ for various direction vectors $\{\boldsymbol{\phi_m}\}$, then we can use a Radon inversion method to reconstruct the joint PMF tensor. This was first shown in \cite{OSullivan1993} for 2D PDFs/PMFs and later in \cite{Kolouri2018}, \cite{Webber2019}. For PDFs with exponentially decreasing Fourier transforms, the Radon-based technique has superior convergence rates compared to approaches like kernel density estimation (KDE) \cite{parzenkde}, \cite{jiangicml}, \cite{Cavalier2000}. 
\vspace{-0.3cm}
\section{Problem Statement and Algorithm}
\label{sec:prob_stmt}
Consider $N$ discrete RVs $\{X_i\}_{i=1}^{i=N}$ each existing in $I$ different states\footnote{This work can be easily  to the case where different RVs had different number of states. Our work can also handle continuous RVs - see Sec. \ref{sec:results}.}. Our aim is to estimate the joint PMF $p(X_1, X_2,.., X_N)$ where $p(.)$ is a low-rank tensor which follows the CPD from Eqn. \ref{eq:tens_dec}. However, recovering the tensor $p$ from its Radon projections taken in general directions $\{\boldsymbol{\phi}_m \in \mathbb{R}^N\}$ is an immensely costly operation. Hence we consider only sparse direction vectors $\{\boldsymbol{\phi}_m\}$ with just two non-zero entries (i.e. we consider linear combinations of just two RVs at a time), which facilitates speedy FBP implementation. Define the set $\mathcal{B} \triangleq \{(j,k):1 \leq j < k \leq N\}$. For $M$ random vectors $\boldsymbol{\phi}_{m} \in \mathbb{R}^2, 1\leq m\leq M$, and for $\boldsymbol{X_{j,k}} \triangleq [X_j, X_k]^t$ for $(j,k)\in \mathcal{B}$, we obtain $M|\mathcal{B}|$ random projections of the form $\boldsymbol{\phi}_{m}^t \boldsymbol{X_{j,k}} \in \mathbb{R}$. We wish to recover the PMF tensor from the 1D PMFs of the form $p(\boldsymbol{\phi}_{m}^t \boldsymbol{X_{j,k}})$. Define $\boldsymbol{Z}_{j,k}$ as the joint probability $p(X_j,X_k)$ for the two RVs $X_j$ and $X_k$. On fixing $j$ and $k$ and stacking $M$ one-dimensional PMFs $p(\boldsymbol{\phi}_{m}^t \boldsymbol{X_{j,k}})$ row-wise, we obtain a matrix $\boldsymbol{Y}_{j,k} \in \mathbb{R}^{M \times I}$. In case of infinite samples, from Eqn. \ref{eq:radon2}:
\begin{equation}
\boldsymbol{Y}_{j,k} = \mathfrak{R}(\boldsymbol{Z}_{j,k}) = \mathfrak{R}(\boldsymbol{A_j}D(\boldsymbol{\lambda})\boldsymbol{A_k}^T),
\label{eq:J}
\end{equation} 
where $D(.)$ is the diagonal operator as defined in Sec. \ref{subsec:3way-2way}, and $\mathfrak{R}$ stands for the Radon transform in multiple directions $\{\boldsymbol{\phi}_m\}_{m=1}^M$.
However the 1D PMFs, and hence each $\boldsymbol{Y}_{j,k}$ can only be \emph{estimated}. For each element in set $\mathcal{B}$, we perform random projections and estimate \emph{one dimensional} PMFs empirically from the data using histogramming, and thus assemble $\boldsymbol{Y}_{j,k}$. The goal is now to determine the underlying mode factors $\boldsymbol{\lambda},\boldsymbol{A_1},\boldsymbol{A_2},...,\boldsymbol{A_N}$ (and thus the joint PMF of $\{X_i\}_{i=1}^N$), which represent these stacked 1D PMF estimates $\{\boldsymbol{Y}_{j,k}\}_{(j,k) \in \mathcal{B}}$ as faithfully as possible. To this end, we formulate the following objective function:
\begin{equation}
J(\{\boldsymbol{A_n}\}_{n=1}^N,\boldsymbol{\lambda}) = \sum_{j,k>j}  \| \boldsymbol{Y}_{j,k} -  \mathfrak{R}(\boldsymbol{A_j}D(\boldsymbol{\lambda})\boldsymbol{A_k}^T)\|_F^2. 
\end{equation}
We cannot directly optimise $J(.)$, say via gradient descent updates, as the mode factors will not be identifiable when $F>I$ as argued in \cite{Ibrahim2020}. To circumvent, we introduce an auxiliary variable $\boldsymbol{Z}_{j,k}$ with the constraint $\boldsymbol{Z}_{j,k} = \boldsymbol{A_j}D(\boldsymbol{\lambda})\boldsymbol{A_k}^T$ and transform it into an unconstrained problem by adding a penalty term, namely
\begin{eqnarray}
J_1(\{\boldsymbol{A_n}\}_{n=1}^N,\boldsymbol{\lambda},\{\boldsymbol{Z}_{j,k}\}) \triangleq \sum_{j,k>j}  \| \boldsymbol{Y}_{j,k} -  \mathfrak{R}(\boldsymbol{Z}_{j,k})\|_F^2 + \nonumber \\ 
\rho\|\boldsymbol{Z}_{j,k} - \boldsymbol{A_j}D(\boldsymbol{\lambda})\boldsymbol{A_k}^T\|_F^2. 
\label{eq:penalty}
\end{eqnarray}
We perform the following three steps to obtain identifiable mode factors: (i) setting the hyper-parameter $\rho = 0$, compute $\boldsymbol{Z}_{j,k}$ from the 1D densities $\boldsymbol{Y}_{j,k}$ using least-squares, (ii) construct $\boldsymbol{\widetilde{Z}}$ from $\boldsymbol{Z}_{j,k}$ as explained in Sec. \ref{subsec:3way-2way}, and (iii) determine the matrices $\boldsymbol{W}$ and $\boldsymbol{H}$ as outputs of the function $\textsc{Spa}(.)$ which are the factors of $\boldsymbol{\widetilde{Z}}$. The mode factors $\{\boldsymbol{A_j}\}$ which are sub-matrices of $\boldsymbol{W}$ and $\boldsymbol{H}$ will now be identifiable as shown in \cite{Ibrahim2020}. We further refine our estimates by choosing $\rho$ via cross-validation, and updating the mode factors and $\boldsymbol{Z_{j,k}}$ till convergence as described in Alg.~\ref{alg:alg1}. The resultant mode factors obtained from optimizing $J_1(.)$ can then be used as a good initial condition for a projected gradient descent on the original objective function $J(.)$ with adaptive step-size.\eat{ For Alg. \ref{alg:alg1}, we assume that we have an access to a function $\textsc{Spa}(.)$ which takes in $\boldsymbol{\widetilde{Z}}$ as input and outputs $\boldsymbol{W, H}$.} In Alg. \ref{alg:alg1}, the operators $\mathfrak{R}$ and $\mathfrak{R}^T$ are implemented as function handles\eat{ using the \texttt{radon} and \texttt{iradon} functions of MATLAB respectively}. \emph{It is important to note that a procedure which stops at the end of step 4 in Alg. \ref{alg:alg1} (equivalent to FBP to obtaining $\boldsymbol{Z}_{j,k}$ from $\boldsymbol{Y}_{j,k}$, followed by the algorithm from \cite{Ibrahim2020}), would necessarily ignore the fact that the 2-way marginals $\boldsymbol{Z}_{j,k}$ are inter-dependent due to common mode factors}. This motivates the further steps in Alg. \ref{alg:alg1} which account for such dependencies, and \emph{the empirical results from Sec. \ref{sec:results} further support their inclusion}. We additionally note that the cost function $J_1(.)$ in Eqn. \ref{eq:penalty} can be easily modified to include further prior information about the density such as it being piece-wise flat (true for PMFs) or smooth. Our algorithm can be viewed as 3 sequential processing blocks which refine the estimates produced by the previous steps in the pipeline. The three blocks are: $\textsf{G1}$ (lines 2-4), $\textsf{G2}$ (lines 6-13), and $\textsf{G3}$ (lines 15-20). We refer to our method as \textsc{Juror}: \textbf{J}oint distribution recovery
\textbf{U}sing \textbf{R}andom projections to \textbf{O}ne dimensional \textbf{R}egion.
\begin{algorithm}[H]
\caption{Recovering Mode Latent Factors from 1D\\ Densities of Random Projections}\label{alg:A5}
\begin{algorithmic}[1]
\Procedure{Juror}{}
\State Set $\{\boldsymbol{Z}_{j,k}\}_{(j,k) \in B}$ to be equal to  arg$\min  \sum_{j,k>j}  \|\boldsymbol{Y}_{j,k} - \mathfrak{R}(\boldsymbol{Z}_{j,k})\|_F^2$
\State Assemble $\boldsymbol{\widetilde{Z}}$ using  $\{\boldsymbol{Z}_{j,k}\}_{(j,k) \in B}$ 
\State ${\boldsymbol{W}^{(0)}, \boldsymbol{H}^{(0)}} \gets \textsc{Spa}(\boldsymbol{\widetilde{Z}})$
\State $\text{converged} \gets \text{False}$, $\rho \gets \rho_0$, $q \gets 1$
\While{$\text{converged} == \text{False}$}
\State Fetch $\{\boldsymbol{A}_n\}_{n=1}^{n=N},\boldsymbol{\lambda}$ from $\boldsymbol{W}^{(q-1)}, \boldsymbol{H}^{(q-1)}$\vspace{0.1cm}
\State ${\boldsymbol{Z}^{(n)}_{j,k}}\hspace{-5pt} = \hspace{-3pt} {(\mathfrak{R}^T\mathfrak{R}+\rho I)^{-1}}$\hspace{-5pt} ${(\mathfrak{R}^T\boldsymbol{Y}_{j,k} + \rho \boldsymbol{A}_jD(\boldsymbol{\lambda})\boldsymbol{A}_k^T)}$
\State Assemble $\boldsymbol{\widetilde{Z}}$ using  $\{{{\boldsymbol{Z}_{j,k}^{(n)}}\}}_{(j,k) \in B}$
\State ${\boldsymbol{W}^{(q)}, \boldsymbol{H}^{(q)}} \gets \textsc{Spa}({\boldsymbol{\widetilde{Z}}})$
\State $q=q+1$
\If {$J_1(.) < \epsilon$} $\text{converged}\gets \text{True}$ \EndIf
\EndWhile
\State Fetch $\{\boldsymbol{A}_n^0\}_{n=1}^{n=N},\boldsymbol{\lambda}^0$ from $\boldsymbol{W}^{(q)}, \boldsymbol{H}^{(q)}$
\State converged$\gets$ False, $q \gets 1$
\While{$\text{converged} == \text{False}$}
\For{$k$ in $1$ to $N$}
\State $\boldsymbol{A}_k^{(q)} \hspace{-2pt} \gets \text{ProjectOnSimplex}(\boldsymbol{A}_k^{(q-1)} - \eta_q\frac{\partial J_1}{\partial \boldsymbol{A}_k}) $
\EndFor
\State $\boldsymbol{\lambda}^{(q)} \gets \text{ProjectOnSimplex}(\boldsymbol{\lambda}^{(q-1)} - \eta_q\frac{\partial J_1}{\partial \boldsymbol{\lambda}}) $
\State $q=q+1$
 \If {$J(.) < \epsilon$} $\text{converged}\gets \text{True}$ \EndIf
\EndWhile
\State \Return $\{\boldsymbol{A_n}\}_{n=1}^{n=N},\boldsymbol{\lambda}$
\item[]
\EndProcedure
\end{algorithmic}
\label{alg:alg1}
\end{algorithm}

\section{Numerical Results}
\label{sec:results}
In this section, we present several PMF estimation results on both synthetic and real-world datasets. For synthetic data, we present results for both discrete and continuous RVs. The synthetic data are created from their mode factors, in the following manner: (1) For PMFs of discrete RVs are created, each entry in $\boldsymbol{A_i}$ is generated i.i.d. from $\textrm{Uniform}[0,1]$ followed by normalizing the columns to have unit $\ell_1$ norm. Elements of $\boldsymbol{\lambda}$ are generated in the same manner. (2) For continuous RVs, we consider cumulative interval measures (CIMs) instead of PDFs and represent them as tensors. For the CIM, each column of $\boldsymbol{A_i}$ is generated by sampling from sinusoidal waves of varying amplitudes, phase and frequency, at regular intervals. 
For synthetic data, $F$ is known to the algorithm before hand and it tries to find the mode factors given just the data sampled from the underlying distribution. For all the experiments, the number of random projections $M = 200$ and $|\mathcal{B}| = {N \choose 2}$. Let $\boldsymbol{\lambda}, \{\boldsymbol{A_n}\}_{n=1}^N$ and $\widehat{\boldsymbol{\lambda}}, \{\widehat{\boldsymbol{A_n}}\}_{n=1}^N$ be the mode factors of the true and estimated PMFs/CIMs respectively. Let $\kappa \in (0,1]$ be the fraction of observed (as opposed to missing) entries in the data-points. Since storing the tensor with $I^N$ entries may be infeasible, we compute the mean squared estimation error in terms of the mode factors: $\textrm{MSE}  \triangleq \sum_{i=1}^{i=N}\frac{\| \widehat{\boldsymbol{ A}_i} - \boldsymbol{A}_i  \|_F^2}{N} + \|\widehat{\boldsymbol{\lambda}} - \boldsymbol{\lambda} \|_2^2$ (as is the norm in \cite{Ibrahim2020,Kargas2018}).
The results are reported in Tables \ref{tab:synthetic1} and \ref{tab:synthetic2} for different number of samples (denoted by $N_s$) for PMF/CIM estimation. In both cases $F > I$ and $\kappa = 1$. The methods compared are (1) \textsc{Ctf} using 3-way marginals from \cite{Kargas2018}; (2) the \textsc{Spa} method using 2-way marginals from \cite{Ibrahim2020}; (3) our technique from Alg. 1 termed \textsc{Juror} with various combinations of the stages of the algorithm defined as \textsc{A}: $\textsf{G1}+\textsf{G2}+\textsf{G3}$, \textsc{B}: $\textsf{G1}+\textsf{G3}$, \textsc{C}: $\textsf{G1}+\textsf{G2}$; and (4) the EM technique from \cite{Yeredor2019MaximumLE} with random initial conditions referred to as \textsc{Rand-em}. In Table 3 we consider the real-world scenario where not all features are observed for every data-point and present results for $\kappa=0.8$ as the probability of observing each feature of a sample.\eat{ In many real-world applications not all features are observed for every data-point. PMF/CIM estimation from marginals is particularly useful in these scenarios, because simple histogramming will discard samples with missing data. In Table \ref{tab:synthetic_missing1} we present results with $\kappa=0.8$ as the probability of observing each feature of a sample.}
\begin{table}[H]
    \centering
    \begin{tabular}{c c c c c c}
    $N_s$ & 100 & 1000 & 5000 & 10000 & 50000 \\ \midrule
   
   \textsc{Ctf} & 0.339 & 0.294 & 0.233 & 0.172 & 0.103\\
   \textsc{Spa}   & 0.295 & 0.264 & 0.196 & 0.143 & 0.084 \\
   \textsc{Juror-a}    & \textbf{0.253} & \textbf{0.205} & \textbf{0.174} & 0.138 & 0.092  \\
     \textsc{Juror-b} & 0.267 & 0.229 & 0.182 &\textbf{ 0.131} & \textbf{0.087} \\
     \textsc{Juror-c}  & 0.262 & 0.217 & 0.185 & 0.140 & 0.098 \\

   \textsc{Rand-em} & 0.284 & 0.246 & 0.204 & 0.149 & 0.106\\ \midrule

    \end{tabular}
    \caption{MSE for PMFs for $F = 25, I=10, N=6, \kappa = 1$}
    \label{tab:synthetic1}
\end{table}
\vspace{-0.5cm}
We also study the absolute tensor errors in addition to MSE in mode factors when storing the tensor is feasible. If $\mathcal{Z}$ is the original tensor and $\mathcal{\hat{Z}}$ is the estimate of the joint distribution tensor, then we measure the mean absolute error defined as MAE $\triangleq \frac{\|\mathcal{\hat{Z}} - \mathcal{Z}\|_F^2}{\|\mathcal{Z}\|_F^2}$. The results for MAE comparisions are shown in Table \ref{tab:mae}.
\vspace{-0.5cm}
\begin{table}[H]
    \centering
    \begin{tabular}{c c c c c }
    $N_s$ & 100 & 1000 & 5000 & 10000  \\ \midrule
   
   \textsc{Ctf} & 0.215 & 0.181 & 0.123 & 0.096 \\
   \textsc{Spa}   & 0.175 & 0.151 & 0.114 & 0.063 \\
   \textsc{Juror-a}    & \textbf{0.158} &\textbf{ 0.136} & \textbf{0.093} & 0.054   \\
     \textsc{Juror-b} & 0.169 & 0.148 & 0.102 &\textbf{ 0.048} \\
     \textsc{Juror-c}  & 0.162 & 0.139 & 0.108 & 0.059  \\

   \textsc{Rand-em} & 0.183 & 0.174 & 0.116 & 0.082 \\ \midrule

    \end{tabular}
    \caption{MAE for PMFs for $F = 15, I=10, N=4, \kappa = 1$}
    \label{tab:mae}
\end{table}

The MSE results in Tables \ref{tab:synthetic1}, \ref{tab:mae}, \ref{tab:synthetic2},  \ref{tab:synthetic_missing1} are obtained after averaging over 5 random runs, and with $M = 200$ Radon projections per pair of RVs. As expected, our algorithm performs better in the low-sample regime where estimating 1D marginals is much more reliable than higher-D marginals. All the methods start converging in the high-sample regime where out method isn't far from the lowest MSE. The results for high-D PMFs with $N=15$ are presented in Table \ref{tab:synthetic_highD}. From the above tests it is evident that $\textsf{G2}$ of the algorithm is crucial when we have less samples and the final step $\textsf{G3}$ is more effective in the high-sample regime. We also study the effect of number of projections $M$ on density estimation using the algorithm \textsc{Spa-r-a}. The results are shown in Table \ref{tab:M_vary}.

\begin{table}[H]
    \centering
    \begin{tabular}{c c c c c c}
   $N_s$  & 1000 & 5000 & 10000 & 50000 & 100000\\ \midrule
   
   \textsc{Ctf}  & 0.318 & 0.289 & 0.227 & 0.191 & 0.164\\
   \textsc{Spa}    & 0.296 & 0.274 & 0.238 & 0.182 & 0.158 \\
   \textsc{Juror-a}     & \textbf{0.271} &\textbf{ 0.259} & \textbf{0.194} & \textbf{0.162 }& 0.149 \\
    \textsc{Juror-b}  & 0.283 & 0.264 & 0.209 & 0.175 & \textbf{0.137 }\\
   \textsc{Juror-b} & 0.276 & 0.268 & 0.204 & 0.179 & 0.155 \\

   \textsc{Rand-em}  & 0.281 & 0.268 & 0.198 & 0.187 & 0.157\\ \midrule
    \end{tabular}
\caption{MSE for CIMs for $F = 90, I = 50, N = 5, \kappa = 1$}
    \label{tab:synthetic2}
\end{table}
\vspace{-0.4cm}
\begin{table}[H]
    \centering
    \begin{tabular}{c c c c c }
     $N_s$ & 100 & 1000 & 10000 & 50000  \\ \midrule
   
   \textsc{Ctf}  & 0.316 & 0.151 & 0.142 & \textbf{0.092}\\
   \textsc{Spa}  & 0.238 & 0.162 & 0.126 & 0.117\\
   \textsc{Juror-a}      & \textbf{0.205} & \textbf{0.147} & 0.131 & 0.103  \\
   
      \textsc{Juror-b}      & 0.225 & 0.159  & \textbf{0.118}  & 0.107   \\
      \textsc{Juror-c}      & 0.218 & 0.154  & 0.126  & 0.114   \\

   \textsc{Rand-em}  & 0.269 & 0.173 & 0.152 & 0.136\\ \midrule
    \end{tabular}
    \caption{MSE for PMFs for $F = 20$, $I = 15$, $N = 5$, $\kappa = 0.8$}
    \label{tab:synthetic_missing1}
\end{table}
\vspace{-0.2cm}

In real-world applications where the true underlying PMF is unknown, we compared the different methods in terms of classification accuracy computed after estimating the joint PMFs of the form $p(\boldsymbol{x},y)$, where $y$ is the label associated with data-point $\boldsymbol{x} \in \mathbb{R}^N$. The classification is done by assigning class label $\hat{y} = \textrm{argmax}_y p(y|\boldsymbol{x}) = \textrm{argmax}_y p(\boldsymbol{x},y)/p(\boldsymbol{x})$ to $\boldsymbol{x}$. We test our algorithms against some common classification methods on two datasets from the UCI repository\footnote{\texttt{https://archive.ics.uci.edu/ml/datasets.php}} - the Car (6D) and the Mushroom (22D) datasets. The value of $F$ was set by cross-validating with the accuracy obtained on the validation split. Commonly used discriminative classifiers such as \textsc{Svm-Rbf} and neural networks, were trained using MATLAB's classification toolbox \cite{Ballabio_2013}. The train-val-test split was 70-10-20 for the car dataset and 50-20-30 for the mushroom dataset. As seen in Table \ref{tab:classification}, our method \textsc{Juror} outperforms other techniques.
\begin{table}[H]
    \centering
    \begin{tabular}{c c c c c c}
     $N_s$ & 1000 & 5000 & 10000 & 50000 & 100000\\ \midrule
   
   \textsc{Ctf} & 0.268 & 0.242 & 0.201 & \textbf{0.154} & \textbf{0.104}\\
   \textsc{Spa}   & 0.245 & 0.237 & 0.226 & 0.176 & 0.126 \\
   \textsc{Juror-a}     & \textbf{0.229} & \textbf{0.208} & \textbf{0.184} & 0.162 & 0.115 \\
      \textsc{Juror-b}     & 0.258 & 0.217  & 0.205  & 0.158  & 0.109  \\
      \textsc{Juror-c}     & 0.241 & 0.225  & 0.197  & 0.171  & 0.119  \\

   \textsc{Rand-em}  & 0.248 & 0.226 & 0.217 & 0.197 & 0.143\\ \midrule
    \end{tabular}
    \caption{MSE for PMFs for $F = 20$, $I = 10$, $N = 15$, $\kappa = 1$}
    \label{tab:synthetic_highD}
\end{table}
\vspace{-0.4cm}
\begin{table}[H]
    \centering
    \begin{tabular}{c c c c c c}
     $M(\downarrow)$ & 100 & 1000 & 5000 & 10000 & 50000 \\ \midrule
   
   200     & 0.253 & 0.205 & 0.174 & 0.138 & 0.092  \\
   150     & 0.264 & 0.214 & 0.188 & 0.151 & 0.106  \\
   100     & 0.289 & 0.267 & 0.223 & 0.184 & 0.129  \\
   50     & 0.384 & 0.357 & 0.305 & 0.264 & 0.213  \\ \midrule

    \end{tabular}
    \caption{MSE for PMFs for $F = 25, I = 10, N = 6, \kappa = 1$}
    \label{tab:M_vary}
\end{table}
\vspace{-0.4cm}
\begin{table}[H]
    \centering 
    \begin{tabular}{ c  c  c } \midrule
     \textbf{Algorithm} & \textbf{Car} & \textbf{Mushroom } \\ \midrule
   
   \textsc{Ctf} & 84.92 & 95.13 \\
   \textsc{Spa}   & 86.45 & 96.01  \\
   \textsc{Juror-a}     & 87.59 &\textbf{ 96.72}  \\
      \textsc{Juror-b}     & 86.37 &95.12  \\
   \textsc{Juror-c}     & \textbf{88.38} &95.79  \\

   \textsc{Rand-em}  & 82.68 & 94.65 \\ 
   \textsc{Logistic Regression} & 82.37 & 95.86\\
   \textsc{Svm-rbf} & 78.32 & 95.74\\
   \textsc{Naive Bayes} & 84.39 & 89.67\\
   \textsc{Neural Net} &85.16 &96.37\\\hline
    \end{tabular}
    \caption{Classification Accuracies on real-world datasets}
    \label{tab:classification}
\end{table}
\section{Conclusion and Future Work}
Combining ideas from tomography and low-rank tensors, we presented a novel algorithm to recover the joint PMF from one-way marginals obtained from random projections of the data. Empirical results suggest that our density estimation method is particularly useful in the low-sample regime where most other estimators under-perform. Some future work may include a theoretical analysis of the proposed method from the point of view of sample complexity and estimating continuous PDFs directly using appropriate basis functions for the mode latent factors.
\bibliographystyle{IEEEbib}
\bibliography{refs}

\end{document}